\documentclass[sigconf]{acmart}
\AtBeginDocument{%
  }
\usepackage{makecell}
\usepackage{booktabs}
\usepackage{pifont}
\usepackage{xcolor}
\usepackage{amsmath}
\usepackage{amsthm}
\usepackage{amsfonts}
\usepackage{multirow}
\usepackage{multicol}
\usepackage{array}     
\newcommand{\xmark}{\textcolor{red}{\ding{55}}}
\newcommand{\cmark}{\textcolor{green}{\ding{51}}}

\usepackage[table,xcdraw]{xcolor}
\usepackage{tcolorbox}

\copyrightyear{2025}
\acmYear{2025}
\setcopyright{cc}
\setcctype{by}
\acmConference[MM '25]{Proceedings of the 33rd ACM International Conference on Multimedia}{October 27--31, 2025}{Dublin, Ireland}
\acmBooktitle{Proceedings of the 33rd ACM International Conference on Multimedia (MM '25), October 27--31, 2025, Dublin, Ireland}\acmDOI{10.1145/3746027.3755744}
\acmISBN{979-8-4007-2035-2/2025/10}

\begin{document}

\title{Reasoning Like Experts: Leveraging Multimodal Large Language Models for Drawing-based Psychoanalysis}

\author{Xueqi Ma}
\authornote{Both authors contributed equally to this research.}
\affiliation{
  \institution{The University of Melbourne}
  \city{Melbourne}
  \country{Australia}}
\email{xueqim@student.unimelb.edu.au}

\author{Yanbei Jiang}
\authornotemark[1]
\affiliation{
  \institution{The University of Melbourne}
  \city{Melbourne}
  \country{Australia}}
\email{yanbeij@student.unimelb.edu.au}

\author{Sarah Erfani}
\affiliation{
  \institution{The University of Melbourne}
  \city{Melbourne}
  \country{Australia}}
\email{sarah.erfani@unimelb.edu.au}

\author{James Bailey}
\affiliation{
  \institution{The University of Melbourne}
  \city{Melbourne}
  \country{Australia}}
\email{baileyj@unimelb.edu.au}

\author{Weifeng Liu}
\affiliation{
  \institution{China University of Petroleum (East China)}
  \city{Qingdao}
  \country{China}}
\email{liuwf@upc.edu.cn}

\author{Krista A. Ehinger}
\affiliation{
  \institution{The University of Melbourne}
  \city{Melbourne}
  \country{Australia}}
\email{kris.ehinger@unimelb.edu.au}

\author{Jey Han Lau}
\affiliation{
  \institution{The University of Melbourne}
  \city{Melbourne}
  \country{Australia}}
\email{jeyhan.lau@gmail.com}

\renewcommand{\shortauthors}{Xueqi Ma et al.}

\begin{abstract}
Multimodal Large Language Models (MLLMs) have demonstrated exceptional performance across various objective multimodal perception tasks, yet their application to subjective, emotionally nuanced domains, such as psychological analysis, remains largely unexplored. In this paper, we introduce \textit{PICK}, a multi-step framework designed for Psychoanalytical Image Comprehension through hierarchical analysis and Knowledge injection with MLLMs, specifically focusing on the House-Tree-Person (HTP) Test, a widely used psychological assessment in clinical practice. First, we decompose drawings containing multiple instances into semantically meaningful sub-drawings, constructing a hierarchical representation that captures spatial structure and content across three levels: single-object level, multi-object level, and whole level. Next, we analyze these sub-drawings at each level with a targeted focus, extracting psychological or emotional insights from their visual cues. We also introduce an HTP knowledge base and design a feature extraction module, trained with reinforcement learning, to generate a psychological profile for single-object level analysis. This profile captures both holistic stylistic features and dynamic object-specific features (such as those of the house, tree, or person), correlating them with psychological states. Finally, we integrate these multi-faceted information to produce a well-informed assessment that aligns with expert-level reasoning. Our approach bridges the gap between MLLMs and specialized expert domains, offering a structured and interpretable framework for understanding human mental states through visual expression. Experimental results demonstrate that the proposed PICK significantly enhances the capability of MLLMs in psychological analysis. It is further validated as a general framework through extensions to emotion understanding tasks. Codes are released at https://github.com/YanbeiJiang/PICK.
\end{abstract}



\maketitle

\section{Introduction}


Multimodal large language models (MLLMs) \cite{zhou2024mlvu,gpt4o,zhu2023minigpt,lu2024deepseek,chen2022pali} have achieved remarkable performance across a wide range of objective multimodal perception tasks, such as image captioning \cite{alayrac2022flamingo,zeng2024meacap}, visual question answering \cite{ghosal2023language,li2023blip}, and cross-modal translation \cite{hong2022cogvideo,rombach2022high}. However, their ability to accurately capture and reason with complex and subjective human emotions and psychological states remains limited \cite{yang2024emollm}. We have seen a growing prevalence of mental health issues among both children and adults, and given the shortage of professional psychologists and the inherently subjective nature of psychological assessments, there is a need to develop transparent, automated approaches that can facilitate early detection of psychological disorders.


\begin{figure*}[t]
	\centering
	\includegraphics[width=5.7in]{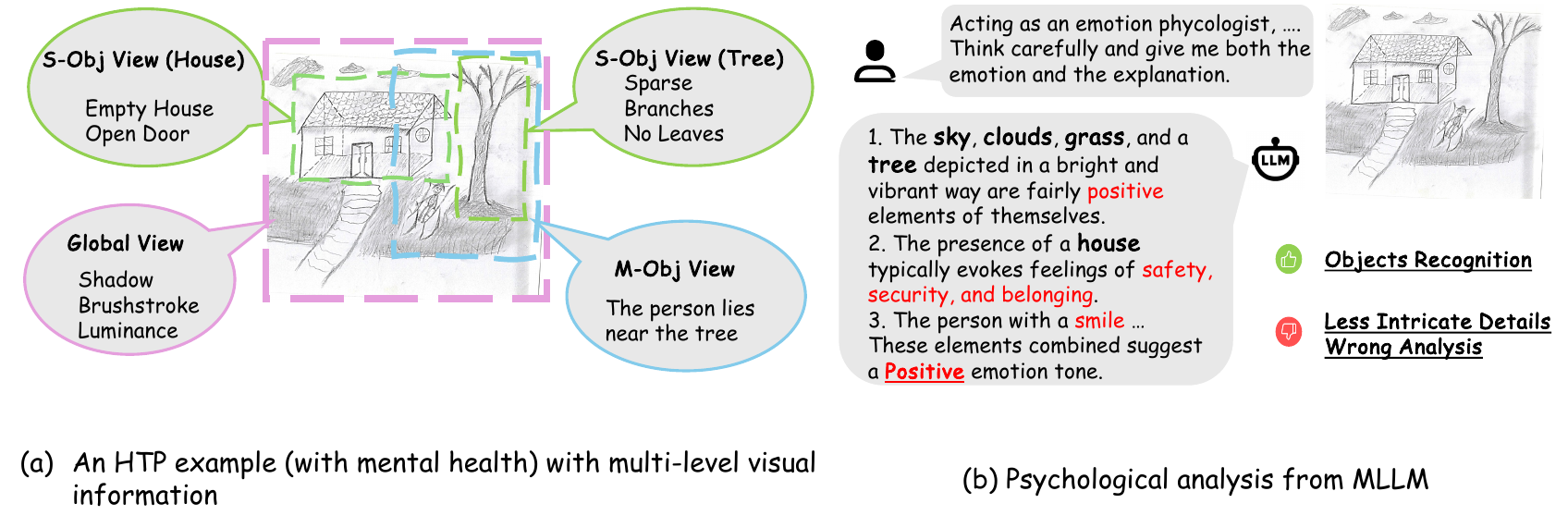}
	\caption{(a) An HTP drawing contains multi-level information related to mental health including single-object (S-Obj), multi-object (M-Obj) and whole levels. Individual elements, such as the house, with attributes like being empty or having an open door, may reflect feelings of loneliness or openness. At the whole level, features like shadowing may suggest isolation. (b) Existing MLLMs can identify key elements (bold text) but struggle to capture the critical information and link it to accurate psychological or mental states (red text).
	}
	\label{fig-example}
\end{figure*}

Among the various methods used in clinical settings, projective tests such as the House-Tree-Person (HTP) test \cite{buck1948htp} have been commonly used to explore an individual's subconscious, mental state, and overall psychological well-being. The HTP test relies on the analysis of drawings created by subjects, based on the premise that the way they depict common objects like houses, trees, and people reflects their underlying emotional states, personality traits, and interpersonal relationships. Therapists analyze visual cues in these drawings, such as size, position, shadow, and detail, to infer about psychological conditions like depression and anxiety \cite{pan2022automated,guo2023analysis}.

However, analyzing HTP drawings with MLLMs presents several challenges: 1) MLLMs, typically pre-trained on natural images, struggle to interpret sketches in a zero-shot manner due to the fundamental differences between sketches and natural images, since sketches are sparser and lack detailed color and texture information; 2) Not all elements in sketches are directly related to psychological or emotional states, as their relevance depends on the specific objects shown, complicating the interpretive process; 3) MLLMs generally lack the subjective expert knowledge needed to accurately connect visual expressions to mental health; and 4) HTP drawings encode information across multiple levels, as shown in Fig. \ref{fig-example} (a): the single-object level (e.g., empty house), the multi-object level which captures spatial and contextual relationships among multiple objects (e.g., the person lies near the tree), and the whole stylistic level (e.g., shadow). Thus, while GPT-4o \cite{gpt4o} can extract key features of houses, trees, people, and other elements in HTP drawings for object classification,  it struggles to capture critical visual cues and accurately correlate them with psychological states, as shown in Fig. \ref{fig-example} (b).



In this paper, we focus on developing an approach for identifying psychologically and emotionally relevant visual features that differentiate individuals with mental health conditions from those without, and forming a psychological assessment. To achieve this, we propose a zero-shot multi-step MLLM-based reasoning framework, \textbf{PICK},
for \textbf{P}sychoanalytical \textbf{I}mage \textbf{C}omprehension through hierarchical analysis with expert \textbf{K}nowledge injection. Since a drawing comprises multiple objects and their interrelationships, we first detect and recognize individual objects, then decompose the drawing into multiple sub-drawings based on object categories. This structured approach enables hierarchical analysis across three levels—Single-Object level, Multi-Object level, and whole level—facilitating a more detailed and interpretable assessment of psychological and emotional states.

Next, we explore the visual information related to psychology or emotion in sub-drawings. Specifically, at the \textbf{single-object level} (e.g., tree), we design an object-tailored prompt paired with a sub-drawing, guiding the MLLM
to capture subtle underlying emotions. This prompt incorporates multiple object-related attributes, including fixed generic features (e.g., size and position) and dynamic object-specific features (e.g., trunk and crown for a tree) generated by a specialized feature extraction module. For the feature extraction module, we construct an HTP Knowledge Base (KB)
to train an emotion-preference reward model, and then fine-tuning a MLLM to generate object-specific features via reinforcement learning using the reward model. This specialized module offers two key benefits: 1) the HTP KB help align the MLLM with expert human knowledge, and 2) the generated dynamic object-specific attributes enable the MLLM to capture subtle visual cues linked to psychological and emotional states.
At the \textbf{multi-object level} (containing multiple key objects) and the \textbf{whole level}, we design prompts that guide MLLMs to analyze inter-object relationships and the overall drawing style, respectively, in relation to the subject's mental state.

Finally, we propose a method to integrate the multiple levels of psycho-visual information
to predict the psychological state (e.g., positive or negative mental state) represented in a drawing.
We demonstrate that our proposed PICK serves as a general framework for multiple zero-shot subjective analysis tasks, including psychological evaluation and emotional understanding.

In summary, our main contributions are as follows:
\begin{itemize}
    \item We present PICK, the first zero-shot framework for understanding psychological and emotional states in drawings by decomposing the image into single-object, multi-object, and whole levels.
    \item We construct an HTP knowledge base and integrate it into a specialized feature extraction module at the single-object level to extract better features that are aligned with expert knowledge.
    \item We conduct extensive experiments on two HTP datasets and two emotion datasets, demonstrating that our approach significantly outperforms baseline models. Notably, it achieves an average F1 score improvement of over 10\% in diagnosing psychological disorders on HTP datasets compared to foundation models.
\end{itemize}  

\section{Related Works}

\subsection{Multimodal Large Language Models}
Multimodal Large Language Models (MLLMs) \cite{jin2024chat,lyu2023macaw,han2024onellm} extend traditional LLMs by incorporating visual and audio inputs through techniques such as CLIP \cite{clip} and additional adaptation modules \cite{baevski2020wav2vec,radford2023robust}. These advancements have enabled MLLMs to tackle a wide range of multimodal tasks, including image captioning \cite{bianco2023improving,dzabraev2024vlrm,jiang2024kale}, visual question answering (VQA) (\cite{hu2024bliva,sterner2024few}), and other language-related capabilities (\cite{wu2024q}).

Despite their success in objective multimodal tasks, MLLMs still struggle with subjective reasoning, particularly in understanding complex human mental activity. While several works \cite{yang2024emollm,vaiani2024emotion,liu2024emollms} have made strides in emotional understanding by leveraging both visual and textual cues, they often fall short in capturing deeper psychological states. This limitation becomes particularly evident when interpreting abstract and culturally influenced visual representations, such as hand-drawn sketches used in psychological assessments. Addressing these challenges requires MLLMs to move beyond conventional pattern recognition and develop a more nuanced ability to reason about emotional and psychological contexts.

\subsection{Drawing Psychoanalysis}
Drawing-based psychoanalysis has long been used as a non-verbal method to assess an individual's psychological state, cognitive processes, and emotional well-being. \citet{carter2021children} applied drawing psychoanalysis to find differences between normal and autistic children. The House-Tree-Person (HTP) test \cite{buck1948htp} is a widely used projective drawing test \cite{perticone1998clinical,crusco2013draw,gantt2003formal} that requires individuals to draw a house, a tree, and a person. These objects are chosen because they are universally familiar from an early age, making the task accessible to children and adults alike.
Therapists analyze HTP drawings to assess psychological phenomena, intelligence, and personality traits by considering factors like object proportions, spatial arrangement, line quality, shading, and omissions \cite{buck1948htp}. 
For example, a small or isolated house may suggest detachment or insecurity, while a tree with exaggerated branches and a weak trunk could indicate emotional instability \citep{guo2023analysis}.
Similarly, missing facial features or distorted proportions in the drawn person may reveal insights into self-perception and interpersonal relationships \cite{hammer1958clinical}.

Despite its clinical significance, traditional HTP analysis remains subjective, heavily reliant on expert interpretation, which can vary across practitioners. While recent deep learning models \cite{salar2023artificial,lee2024generating,lin2022house,pan2022automated,xie2023interpretable,zhang2024feasibility} have demonstrated the ability to classify different categories of houses, trees, and people, they still lack the depth of reasoning required for psychological assessments. 
 MLLMs have been recently explored to assist in feature extraction and analysis in HTP problem \cite{xu2025psychological}, but they continue to struggle with understanding detailed visual cues and the implicit psychological meanings embedded in drawings.

\begin{figure*}[]
	\centering
	\includegraphics[width=6.2in]{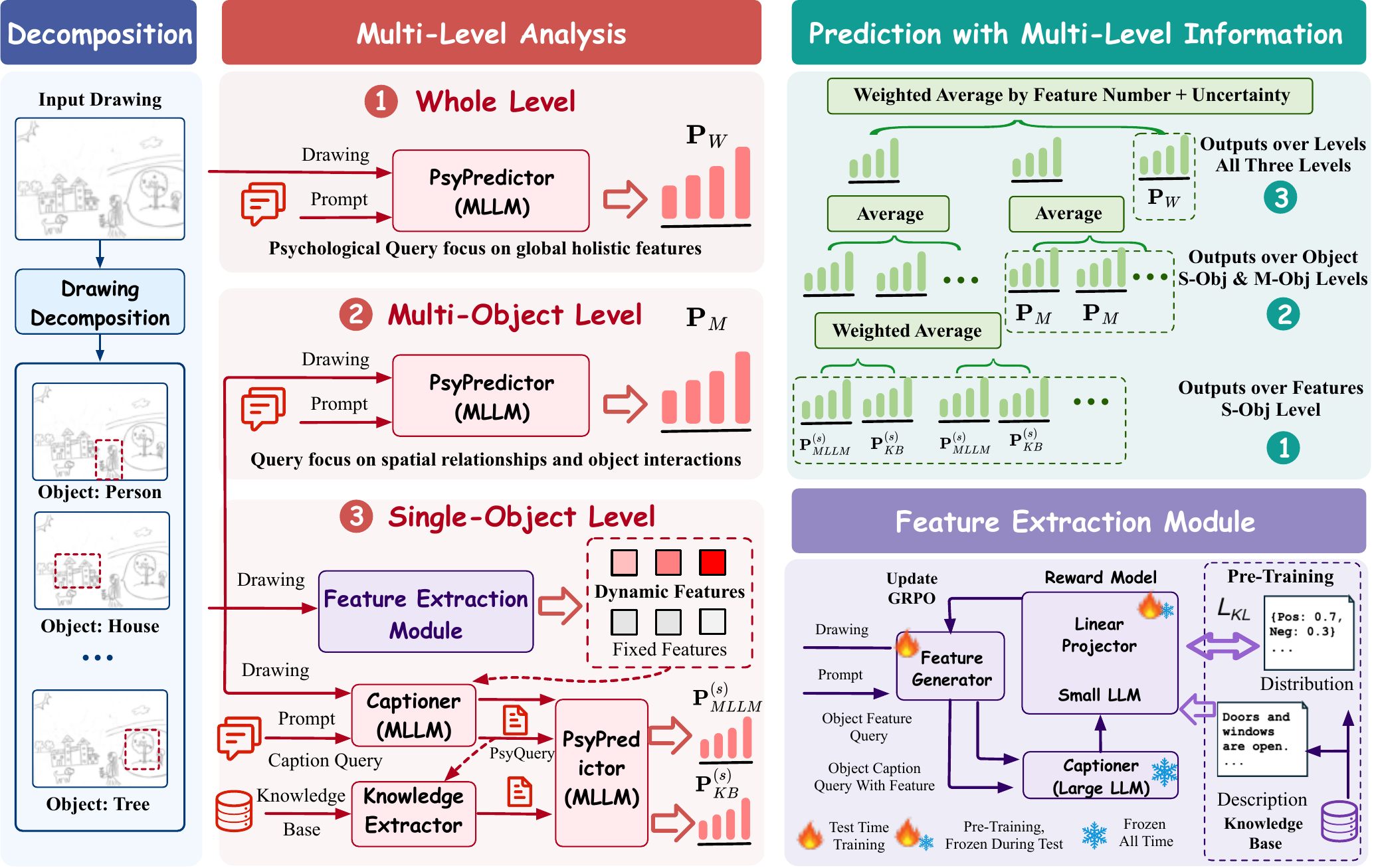}
	\caption{The framework of PICK, including (1) Input Decomposition: The input drawing is divided into multiple sub-drawings.
(2) Multi-Level Analysis: This includes tailored prompts for whole level and multi-object level analysis, a specialized feature extraction module, and the integration of a knowledge base (KB) for single-object level analysis.
(3) Combined Prediction: Multi-level information is aggregated to produce a final prediction.
	}
	\label{fig-frame}
\end{figure*}

\section{PICK}
In this paper, we propose PICK, a multi-step reasoning framework that emulates expert observation and reasoning (illustrated in Figure \ref{fig-frame}). First, we decompose a drawing into semantically meaningful sub-drawings based on object categories, enabling hierarchical analysis at the single-object, multi-object, and whole-drawing levels. We then extract psychological and emotional insights from these levels.  At the single-object level, we introduce a feature extractor module to generate dynamic, object-specific features, which are then combined with fixed generic features to construct an single object level psychological profile. Additionally, we extract expert knowledge from the HTP knowledge base (KB) to further enhance single-object level analysis and interpretation. Finally, we integrate information across multiple levels to ultimately formulate a psychological prediction, specifically a binary classification of positive and negative mental states.

\subsection{Hierarchical Drawing Decomposition}

In hierarchical drawing decomposition, we aim to extract both structural spatial information and semantic information by breaking the drawing down into meaningful components.
For the full drawing, we utilize GroundingDINO \cite{liu2024grounding}, an open-set object detector, to identify and classify main objects (i.e., houses, trees, people), and other elements (e.g., sun, flower, etc). For each main object, we generate an single-object level sub-drawing by retaining the original image and use its detected bounding box to highlight the focused region.
Due to spatial proximity, the bounding boxes of certain neighboring elements may overlap with the bounding box of the main object, potentially influencing its interpretation (e.g., flowers near a tree may evoke a sense of happiness, with the tree as the main object and the flowers as neighboring elements). Therefore, these relevant neighboring elements are also included in the focused region of the single-object sub-drawing.

To account for interrelationships among objects, we merge the regions of multiple surrounding main objects with similar spatial and semantic relevance into a larger bounding box, forming multiple-object level sub-drawings. This larger bounding box is created by extending the edges of the bounding boxes of all included objects.
Notably, single-object sub-drawings contain only one main object, while multiple-object sub-drawings encompass multiple main objects. 
At the multiple-object level, if two sub-drawings overlap by more than 90\%, we discard one of them.
Simultaneously, we analyze the entire drawing for the whole-level assessment. Assume there are $N$ ($N >3$ when multiple houses, trees, or people are present) main objects in a drawing, finally we obtain $N$ single-object level sub-drawings, $M$ multiple-object level sub-drawings, and $1$ whole level drawing.

\subsection{Single-Object Level Analysis}
What specific features of these different objects can reveal a subject’s underlying psychological state and how can we guide MLLMs to generate accurate analysis based on these detailed visual cues? 
Therapists typically analyze the objects according to \cite{lee2024generating,pan2022automated} ---
\textbf{generic features}: factors such as size, position, shading, stroke intensity, composition, etc; and
\textbf{object-specific details}: features such as a broken house, a locked door, or an unusually large tree crown.

While generic features are relatively fixed, with a clear association to psychological states (e.g., individuals with depression often use more black and shadows in their drawings compared to those without \cite{buck1948htp,barrick2002color}), object-specific details are much more dynamic and complex \cite{guo2023analysis}. That is, each object possesses a unique set of attributes and the same object may appear significantly different across various sketches, and the relationship between object attributes and psychological states is highly context-dependent. To address these challenges, we design a feature extraction module that extracts dynamic, object-specific features from input sub-drawings. We then craft prompts incorporating these dynamic features alongside predefined fixed features to guide the MLLM in focusing on the most relevant aspects of the sketches.



\subsubsection{Object-Specific Features Generation by a Fine-tuned MLLM}
\label{obj}

The feature extraction module (shown in Figure \ref{fig-frame}) is a specialized component designed to extract dynamic object-specific features. Specifically, we train an emotion-preference reward model using data from the HTP Knowledge Base and employ it to fine-tune a feature generator—a lightweight MLLM—denoted as $\mathcal{M}_g$, via reinforcement learning. This process enables the generator to produce more psychologically relevant object-specific features.


\paragraph{\textbf{Emotion-Preference Reward Model}}

The emotion-preference reward model is developed to evaluate the intensity of emotions in text and assigns a preference score based on their strength, with higher scores indicating stronger emotions.

To pre-train this reward model for HTP task, we construct a knowledge base called \textbf{HTP KB}, consisting of 4,879 triplets in the format (\text{drawing description}, \text{relation}, \text{inferred mental state}). For example: "Neck is painted in black, indicates, Anxiety". These triplets were manually derived from existing psychological literature \citep{massironi2001psychology,oster2004using} (details in Appendix \ref{app_KB}). To employ this KB dataset for training a reward model, we then use GPT-4o \cite{gpt4o} to generate an emotion distribution, categorizing each triplet into positive and negative proportions. Since the tail entities like "Anxiety" in triplets often explicitly express emotions, we directly use them as the predefined emotion distribution.

The reward model is constructed using a lightweight transformer-based LLM followed by a linear layer. It takes triplet descriptions as input and output a positive and negative soft emotion distribution. The parameters of both the LLM and the linear layer are optimized using KL divergence loss, ensuring alignment with the predefined emotional distribution.
After pre-training, the reward score is calculated as 
${r} = 1 - \frac{H(p)}{\log K}$, which ensures an output range between 0 to 1. 
Here, $K$ is the number of classes, $\log K$ is the uniform distribution, $p$ is the predicted distribution over $K$ classes, $H(p) = -\sum_{k=1}^K p_k \log p_k$ is the entropy of $p$. This formulation assigns higher scores to emotion distributions with lower entropy, indicating stronger and more confident emotional predictions.

\paragraph{\textbf{Object-Specific Features Extraction}}
Using the reward model, we employ a test-time training strategy to adapt the feature generator (a small MLLM) through reinforcement learning (RL), enabling it to extract dynamic object-specific features link to sharper emotion state for each sub-drawing. 

Specifically, for each input sub-drawing, we first prompt the feature generator $\mathcal{M}_g$ to produce detailed, object-specific attributes (e.g., for a tree, it might be "leaves") that are distinct from the existing feature pool, which includes both fixed generic features and previously identified object-specific details. We then design an object caption prompt incorporating the generated feature ("leaves"), and use a Captioner $\mathcal{M}_c$ (a parameter-frozen large MLLM) to generate a descriptive output of the sub-drawing. This description is subsequently fed into the reward model to compute a emotional relevance score, which serves as feedback for refining the model’s feature generation process using Group Relative Policy Optimization (GRPO) \cite{shao2024deepseekmath}. Through this iterative optimization, the $\mathcal{M}_g$ is progressively updated to extract psychologically meaningful object-specific features (e.g., evolving from "leaves" to "large crown") that may reflect underlying mental states, such as loneliness, anxiety, or openness. Note that the $\mathcal{M}_g$ is fine-tuned during test time using test drawings, allowing zero-shot evaluation.

\subsubsection{Prediction for Single-Object Level Sub-drawings}

With the fixed generic features and the generated dynamic object-specific features, we prompt Captioner $\mathcal{M}_c$ to generate a description for each feature of the object. Subsequently, we ask a psychological state predictor (PsyPredictor) $\mathcal{M}_p$ (a parameter-frozen MLLM) to produce a psychological state (or emotion) distribution $P_{\text{MLLM}}^{(S)}$
with a confidence score between 0 and 1 for each description, tailored to the corresponding sub-drawing.

To mitigate potential misinterpretations of subtle mental states in sketches due to insufficient domain-specific training, we incorporate the KB, aligned with human knowledge, as an auxiliary guide. Our knowledge extractor retrieves relevant KB information by matching the embeddings of generated descriptions with the head entities of KB triplets using cosine similarity. The triplets corresponding to the most semantically similar head entities are used to generate a KB-based description. Finally, we prompt the PsyPredictor $\mathcal{M}_p$ with this KB-based description to obtain another psychological or emotional distribution $P_{\text{KB}}^{(S)}$ (see Appendix \ref{app_prompt} for prompt details).

We merge the prediction distributions from the MLLM and the KB to get the combined distribution for dynamic object-specific feature dimension $i$ with

\begin{equation}
p_{i} = \frac{\exp(c_i) \cdot P_{\text{MLLM}_{i}}^{(S)} + \exp(s_i) \cdot P_{\text{KB}_{i}}^{(S)}}{\exp(c_i) + \exp(s_i)},  
\label{MLLM_KB}
\end{equation}
where $P_{\text{MLLM}_{i}}^{(S)}$ is the prediction distribution from the MLLM, $c_i$ is the confidence score from the MLLM (between 0 and 1), $P_{\text{KB}_{i}}^{(S)}$ is the prediction distribution derived from the KB, and $s_i$ is the cosine similarity between the MLLM generated description embedding and the most similar KB description embedding. 
Assume we have $K$ single-object level features, by averaging across outputs of $K$ feature dimensions, we obtain the output $P_{S_i}=\frac{1}{K} \sum_{i=1}^K p_i$ for each sub-drawing $i$ at single-object level. 

\subsection{Multi-Object Level and Whole-Level Analysis}

For each multiple-object level sub-drawing $i$, we directly prompt the PsyPredictor $\mathcal{M}_p$ to generate a multiple-object level mental state distribution $P_{M_i}$ , considering spatial relationships and object interactions within the focused region (see Appendix \ref{app_prompt} for prompt details). This helps uncover implicit psychological cues, such as emotional distance between figures or security conveyed by the environment. 

For the whole-level drawing, we prompt the psychological state predictor to generate the whole-level mental state distribution $P_{W}$, focusing on global holistic features like size, position, shading, stroke consistency, and artistic style (see Appendix \ref{app_prompt} for prompt details). This approach enables the inference of broader psychological patterns and emotional states based on the overall composition and arrangement of elements. 

\subsection{Prediction Based on Multi-level Analysis}



Our hierarchical drawing analysis provides multi-level outputs, each contributing to an understanding of the drawing’s psychological implications. 

At the single-object and multiple-object levels, we compute the final outputs $P_S=\frac{1}{N} \sum_{i=1}^N P_{S_i}$ and $P_M=\frac{1}{M} \sum_{i=1}^M P_{M_i}$ by averaging the predictions of their respective sub-drawings. Since there is only one whole-level drawing, the distribution $P_W$ directly serves as the final whole-level output.

Next, we perform a weighted average across the three levels, where the weights are determined by the information content (the quantity and the uncertainty of soft distributions) of each level:

\begin{equation}
w_l = \frac{n_l \cdot (1 - H(P_l))} {\sum_{l \in \{S, M, W\}} n_l \cdot (1 - H(P_l))}, 
\end{equation}
where $n_l =
\begin{cases} 
N & \text{for } l = S \\
M & \text{for } l = M \\
1 & \text{for } l = W
\end{cases}
$, $ H(p_l) = -\sum_{k=1}^K p_{l,k} \log p_{l,k} $ is the entropy of the averaged distribution $ p_l $ over $k$ classes for that level. Here, $k=2$ for binary classification.
The final distribution of the input drawing is then given by $P_{\text{final}} = \sum_{l \in \{O, S, W\}} w_l \cdot P_l$. 
Finally, the label corresponding to the highest probability in 
$P_{\text{final}}$ is selected as the final prediction. This approach ensures that the most psychologically significant level is prioritized while integrating multi-level insights into a unified decision.



\section{Experiments}
We evaluate the performance of our proposed PICK on two tasks including psychological analysis and emotion understanding. For psychological analysis, we utilize two HTP datasets:
\textbf{HTP\_College} consists of 2,093 HTP drawings collected from the Psychological Testing Center at China University of Petroleum \cite{pan2022automated}, while \textbf{HTP\_Child} includes 257 drawings provided by the Department of Psychology at Istanbul Bilgi University \cite{salar2023artificial}.
For emotion classification task which involves categorizing emotions or feelings from images, we use two widely adopted datasets: \textbf{ArtPhoto} \cite{machajdik2010affective} and \textbf{Emotion6} \cite{emotion6}. Additional details about these datasets can be found in Appendix \ref{app_datasets}.

\begin{table*}[ht]

  \caption{Results on HTP\_College and HTP\_Child(Aggressive) datasets.}
  \centering
  \setlength{\tabcolsep}{5pt}
  \renewcommand{\arraystretch}{1.1}
  \resizebox{0.88\textwidth}{!}{%
  \begin{tabular}{l
      *{7}{c}
      *{7}{c} }
    \toprule
    \multirow{3}{*}{\textbf{Method}} 
      & \multicolumn{7}{c}{\textbf{HTP\_College}} 
      & \multicolumn{7}{c}{\textbf{HTP\_Child(Aggressive)}} \\
    \cmidrule(lr){2-8} \cmidrule(l){9-15}
      & \multirow{2}{*}{Acc} 
      & \multicolumn{3}{c}{Positive} & \multicolumn{3}{c}{Negative} 
      & \multirow{2}{*}{Acc} 
      & \multicolumn{3}{c}{Positive} & \multicolumn{3}{c}{Negative} \\
    \cmidrule(lr){3-5} \cmidrule(lr){6-8} \cmidrule(lr){10-12} \cmidrule(l){13-15}
      &   & F1 & Prec & Rec & F1 & Prec & Rec 
      &   & F1 & Prec & Rec & F1 & Prec & Rec \\
    \midrule
    GPT-4o                & 68.9  & 80.0 & 92.0 & 70.8 & 29.5 & 20.2 & 54.6  & 91.7  & \textbf{95.7} & 91.7 & \textbf{100}  & 0    & 0    & 0    \\
    Gemini-2.0-Flash      & 80.8  & 88.9 & 90.7 & 87.1 & 29.7 & 26.4 & 34.0  & 75.0  & 85.0 & \textbf{94.4} & 77.3 & 25.0 & 16.7 & \textbf{50.0} \\
    Qwen2.5-VL-7B         & 66.9  & 78.2 & \textbf{92.9} & 67.6 & 30.9 & 20.6 & \textbf{62.0}  & 85.4 & 92.0 & 93.0 & 90.9 & 22.2 & 20.0 & 25.0 \\
    Qwen2.5-VL-72B        & 71.3  & 81.8 & 92.5 & 73.4 & 32.0 & 22.3 & 56.4  & 91.7  & 95.7 & 91.7 & \textbf{100}  & 0    & 0    & 0    \\
    Idefics3-8B           & 71.1  & 82.3 & 89.7 & 76.0 & 22.8 & 16.7 & 35.6  & 75.0  & 85.0 & \textbf{94.4} & 77.3 & 25.0 & 16.7 & \textbf{50.0} \\
    InternVL2.5-8B        & 69.0  & 80.5 & 90.1 & 72.8 & 24.1 & 17.1 & 41.2  & 85.4  & 92.0 & 93.0 & 90.9 & 22.2 & 20.0  & 25.0 \\
    InternVL2.5-78B       & 79.6  & 88.1 & 90.8 & 85.6 & 28.5 & 24.3 & 34.6  & 83.3  & 90.7 & 92.9 & 88.6 & 20.0 & 16.7 & 25.0 \\
    \hline
 \rowcolor{gray!20} \textbf{Ours (Based on Gemini)} 
                          & \textbf{84.7}  & \textbf{91.1} & 92.5 & \textbf{89.9} & \textbf{41.7} & \textbf{38.2} & 46.0  & \textbf{91.7}  & 95.6 & 93.5 & 97.7 & \textbf{33.3} & \textbf{50.0} & 25.0 \\
    \bottomrule
  \end{tabular}%
  }
  \label{tab:HTP_college_childAggressive}

\end{table*}

\subsection{Experimental Settings and Evaluation Metrics}

We compare PICK with a range of state-of-the-art MLLM baselines, including both open source (Qwen2.5-VL \cite{bai2025qwen2}, Idefics3 \cite{Idefics2}, and InternVL2.5 \cite{chen2024internvl}) and close source models (GPT-4o \cite{gpt4o} and Gemini-2.0-Flash \cite{balestri2025gender}).
For more details about these models, please refer to Appendix \ref{app_baselines}.

In PICK, the "Captioner" used to generate text descriptions and the "PsyPredictor" for predicting the psychological state or emotion can be any other MLLMs. For the main results in Section \ref{main_results}, we use Gemini-2.0-Flash as both the "Captioner" and "PsyPredictor," and other MLLMs in the ablation study (Section \ref{abl_results}). In Section \ref{obj}, the small LLM used in reward model 
is Qwen2.5 \citep{yang2024qwen2}, while the feature generator in feature extraction module
is Qwen2.5-VL. 
In PICK, we utilize two key generic features (size and position) and two dynamic object-specific features for each input sub-drawing for HTP datasets. Since there are no fixed generic features for emotion analysis, we extract only two dynamic features for the emotion datasets.
Regarding KB, we construct an HTP KB for the HTP datasets, while for the emotion datasets, we introduce an emotion KB from \cite{mohamed2022okay} by aligning emotions (restricted to those present in the datasets) with image captions, resulting in a total of 237,979 triplets.

For HTP\_College, we consider "without any mental health issue" as the positive class and "with mental health issue" as the negative class. For HTP\_Child, there are three mental health conditions: aggressive, anxious, and depressed. In each evaluation, individuals exhibiting a condition are classified as the negative class, while those without it are classified as the positive class. For the emotion datasets, the task is multi-class classification.
Given the imbalanced class distribution (e.g., 1843 positive examples vs. 285 negative examples in HTP\_College) in HTP datasets, we report not only overall accuracy but also the F1 score, precision, and recall for both classes to provide a more comprehensive evaluation of the model's performance. For the two balanced emotion datasets, we compute the macro average across classes, assuming each class holds equal importance.

For emotion datasets, we use the following weighted average across the three levels, where the weights are determined by:
\begin{equation*}
w_l = \frac{ (1 - H(P_l))} {\sum_{l \in \{S, M, W\}} (1 - H(P_l))}.
\end{equation*}
Since an image may contain a large amount of objects and there are no clearly dominant main objects as in the HTP task, we omit the consideration of the number of sub-images in each level.
The final emotion distribution of the input image is then given by $P_{\text{final}} = \sum_{l \in \{O, S, W\}} w_l \cdot P_l$. 
Finally, the label corresponding to the highest probability in 
$P_{\text{final}}$ is selected as the final prediction. 

The feature extraction module of PICK was trained on two NVIDIA A100 80GB GPUs, with an average processing time of approximately 40 seconds per instance. Inference was performed on a single NVIDIA A100 80GB GPU, taking about 50 seconds per instance on average.





\subsection{Main Results}
\label{main_results}

Table \ref{tab:HTP_college_childAggressive} presents the experimental results for all methods on the HTP\_College dataset and the HTP\_Child dataset with the aggressive analysis.
We observe that PICK (based on Gemini) achieves the highest overall accuracy across all datasets and outperforms its backbone, Gemini-2.0-Flash, by significant margins of 3.9\% and 16.7\% on two datasets, respectively.  While the baselines perform well on the positive class, they struggle to detect individuals with mental health issues, sometimes even failing entirely to detect the negative class (i.e., they tend to produce a positive analysis). Our proposed method achieves the best or comparable results on the F1 score for the positive class across all datasets, and surprisingly, we note that PICK demonstrates a 9.7\% improvement over the second-best Qwen2.5-VL-72B on the negative evaluation of HTP\_College. This indicates that PICK is more effective at identifying psychological disorders, which could play a crucial role in early diagnosis and intervention. 
Additionally, on HTP\_Child datasets, our method also surpasses other state-of-the-art large vision-language models in evaluating various mental health conditions, including anxious, and depressed in Table \ref{tab:HTP_childAnxious_childDepressed} in Appendix \ref{app_results}. The superior performance of our method in predicting the mental health of both college students and children highlights the effectiveness of PICK in capturing psychological states from visual cues with expert-level analysis.

\begin{table}[]
  \caption{Results on Emotion datasets.}
  \centering
  \setlength{\tabcolsep}{4pt}
  \renewcommand{\arraystretch}{1.1}
  \resizebox{0.47\textwidth}{!}{%
  \begin{tabular}{lcccccccc}
    \toprule
    \multirow{2}{*}{\textbf{Method}}  
    & \multicolumn{4}{c}{\textbf{Emotion6}} 
    & \multicolumn{4}{c}{\textbf{ArtPhoto}} \\
    \cmidrule(r){2-5} \cmidrule(l){6-9}
    & {Acc} & {F1} & {Prec} & {Rec} 
    & {Acc} & {F1} & {Prec} & {Rec} \\
    \midrule
    GPT-4o & 68.2 & 65.2 & 71.6 & 68.3 & 47.4 & 45.7 & \textbf{58.7} & 45.5 \\
    Gemini-2.0-Flash & 69.8 & 66.0 & 72.5 & 69.9 & 49.6 & 49.8 & 56.3 & 48.7 \\
    Qwen2.5-VL-7B & 68.2 & 62.4 & \textbf{73.5} & 68.2 & 43.9 & 42.7 & 52.6 & 43.3 \\
    Qwen2.5-VL-72B & 67.1 & 62.9 & 69.6 & 67.1 & 42.7 & 43.6 & 53.7 & 42.7 \\
    Idefics3-8B & 63.1 & 57.9 & 73.0 & 63.1 & 44.5 & 41.8 & 49.8 & 41.8 \\
    InternVL2.5-8B & 64.5 & 61.3 & 65.8 & 64.5 & 45.2 & 43.8 & 49.2 & 42.4 \\
    InternVL2.5-78B & 68.2 & 62.9 & 73.4 & 68.2 & 47.4 & 48.2 & 55.9 & 47.4 \\
    \hline
  \rowcolor{gray!20}   \textbf{Ours (Based on Gemini)} & \textbf{70.3} & \textbf{66.4} & \textbf{73.0} & \textbf{70.3} & \textbf{50.1} & \textbf{51.5} & 58.1 & \textbf{50.1} \\
  \bottomrule
  \end{tabular}
  }
  \label{main-emotion}
\end{table}



We next demonstrate the generalization ability of PICK by applying it to emotion classification. Table \ref{main-emotion} compares its performance with state-of-the-art large MLLMs. PICK consistently outperforms baselines across multiple emotion datasets and evaluation metrics, highlighting its effectiveness as a versatile model for subject-domain tasks. PICK excels in capturing subtle differences and intricate relationships between emotional states and visual cues. The results underscore that PICK not only surpasses existing models in accuracy but also excels in generalization, setting a new benchmark for emotion recognition in visual content.

In our experiments, we evaluated models of different sizes, including Qwen2.5-VL-7B vs. 72B and InternVL2.5-8B vs. 78B. As shown in Table \ref{tab:HTP_college_childAggressive}, larger models generally achieved higher overall accuracy on the two HTP datasets. However, for the negative class specifically, the impact of model size was limited—smaller models often performed comparably to, or even slightly better than, their larger counterparts. On the emotion datasets, the two Qwen models showed similar performance, while the larger InternVL2.5 model outperformed the smaller one. These results suggest that increasing model size does not necessarily lead to better performance on subjective or emotionally nuanced tasks.

\definecolor{lightpink}{HTML}{FAEAE1}
\definecolor{iceblue}{HTML}{E0F5FF}
\tcbset{
  pinkbox/.style={
    colback=lightpink,
    colframe=lightpink,
    width = 1.0cm,
    height = 0.35cm,
    halign=center, 
    valign=center, 
  }
}
\tcbset{
  bluebox/.style={
    colback=iceblue,
    colframe=iceblue,
    width = 1.0cm,
    height = 0.35cm,
    halign=center, 
    valign=center, 
  }
}

\begin{table*}[]
    \caption{Comparison of various generative models. Vanilla is the vanilla MLLM. HTP\_Child in here is aggressive analysis. The results are F1 scores in positive F1 / negative F1.}
    \centering
    \setlength{\tabcolsep}{3pt}
    \resizebox{0.75\linewidth}{!}{
    \renewcommand{\arraystretch}{1.0}
    \begin{tabular}{lcccccccc}
        \toprule
        & \multicolumn{2}{c}{HTP\_College} & \multicolumn{2}{c}{HTP\_Child} & \multicolumn{2}{c}{Emotion6} & \multicolumn{2}{c}{ArtPhoto} \\
        \cmidrule(r){2-3} \cmidrule(r){4-5} \cmidrule(r){6-7} \cmidrule(r){8-9}
        Model & Vanilla & Ours & Vanilla & Ours & Vanilla & Ours & Vanilla & Ours \\
        \midrule
        InternVL2.5-8B
        & 80.5/24.1 & 87.0/25.5  
        & 92.0/22.2 & 93.0/40.0  
        & 61.3 & 63.3  
        & 43.8 & 46.9  \\

        Qwen2.5-VL-7B 
        & 78.2/30.9 & 78.9/34.9  
        & 92.0/22.2 & 89.2/30.8  
        & 62.4 & 61.1  
        & 42.7 & 43.2  \\

        GPT-4o
        & 80.0/29.5 & 88.8/40.1 
        & 95.7/0 & 95.7/0  
        & 65.2 & 65.1  
        & 45.7 & 49.1  \\
        
        \bottomrule
    \end{tabular}
    }
    \label{tab:mllm_performance}
\end{table*}


\begin{table*}[t]
  \caption{The ablation study on hierarchical framework. HTP\_Child in here is aggressive analysis. The results are F1 scores, presented in the format: positive F1 / negative F1. M-Obj is multi-object while S-Obj indicates single-object.}
  \centering
  \setlength{\tabcolsep}{2pt}
  \renewcommand{\arraystretch}{1.1}
  \resizebox{0.75\textwidth}{!}{%
    \begin{tabular}{cccccccc}
    \toprule
    Whole & M-Obj & S-Obj & HTP\_College & HTP\_Child & Emotion6 & Artphoto \\
    \midrule
    \cmark & \xmark & \xmark & $90.8_{\tiny \begin{tcolorbox}[bluebox, left=0pt]{\text{$\downarrow$ 0.3}} \end{tcolorbox}}$ / $22.2_{\tiny \begin{tcolorbox}[bluebox, left=0pt]{\text{$\downarrow$ 19.5}} \end{tcolorbox}}$ & $92.0_{\tiny \begin{tcolorbox}[bluebox, left=0pt]{\text{$\downarrow$ 3.6}} \end{tcolorbox}}$ / $22.2_{\tiny \begin{tcolorbox}[bluebox, left=0pt]{\text{$\downarrow$ 11.1}} \end{tcolorbox}}$ & $66.7_{\tiny \begin{tcolorbox}[pinkbox, left=0pt]{\text{$\uparrow$ 0.3}} \end{tcolorbox}}$ &  $51.4_{\tiny \begin{tcolorbox}[bluebox, left=0pt]{\text{$\downarrow$ 0.1}} \end{tcolorbox}}$ \\
    
    \xmark & \cmark & \xmark & $91.5_{\tiny \begin{tcolorbox}[pinkbox, left=0pt]{\text{$\uparrow$ 0.4}} \end{tcolorbox}}$ / $17.8_{\tiny \begin{tcolorbox}[bluebox, left=0pt]{\text{$\downarrow$ 23.9}} \end{tcolorbox}}$ & $95.7_{\tiny \begin{tcolorbox}[pinkbox, left=0pt]{\text{$\uparrow$ 0.1}} \end{tcolorbox}}$ / $00.0_{\tiny \begin{tcolorbox}[bluebox, left=0pt]{\text{$\downarrow$ 33.3}} \end{tcolorbox}}$ &  $66.4_{\tiny \begin{tcolorbox}[bluebox, left=0pt]{\text{0.0}} \end{tcolorbox}}$ & $50.9_{\tiny \begin{tcolorbox}[bluebox, left=0pt]{\text{$\downarrow$ 0.6}} \end{tcolorbox}}$ \\
    
    \xmark & \xmark & \cmark & $81.4_{\tiny \begin{tcolorbox}[bluebox, left=0pt]{\text{$\downarrow$ 9.7}} \end{tcolorbox}}$ / $36.5_{\tiny \begin{tcolorbox}[bluebox, left=0pt]{\text{$\downarrow$ 5.2}} \end{tcolorbox}}$ & $63.6_{\tiny \begin{tcolorbox}[bluebox, left=0pt]{\text{$\downarrow$ 22}} \end{tcolorbox}}$ / $20.0_{\tiny \begin{tcolorbox}[bluebox, left=0pt]{\text{$\downarrow$ 13.3}} \end{tcolorbox}}$ & $57.8_{\tiny \begin{tcolorbox}[bluebox, left=0pt]{\text{$\downarrow$ 8.6}} \end{tcolorbox}}$ & $42.2_{\tiny \begin{tcolorbox}[bluebox, left=0pt]{\text{$\downarrow$ 9.3}} \end{tcolorbox}}$ \\
    
    \xmark & \cmark & \cmark &  $90.7_{\tiny \begin{tcolorbox}[bluebox, left=0pt]{\text{$\downarrow$ 0.4}} \end{tcolorbox}}$ / $44.2_{\tiny \begin{tcolorbox}[pinkbox, left=0pt]{\text{$\uparrow$ 2.5}} \end{tcolorbox}}$ & $89.6_{\tiny \begin{tcolorbox}[bluebox, left=0pt]{\text{$\downarrow$ 6.0}} \end{tcolorbox}}$ / $00.0_{\tiny \begin{tcolorbox}[bluebox, left=0pt]{\text{$\downarrow$ 33.3}} \end{tcolorbox}}$ & $62.9_{\tiny \begin{tcolorbox}[bluebox, left=0pt]{\text{$\downarrow$ 3.5}} \end{tcolorbox}}$ & $45.3_{\tiny \begin{tcolorbox}[bluebox, left=0pt]{\text{$\downarrow$ 6.2}} \end{tcolorbox}}$ \\
    
    \cmark & \xmark & \cmark & $85.9_{\tiny \begin{tcolorbox}[bluebox, left=0pt]{\text{$\downarrow$ 5.2}} \end{tcolorbox}}$ / $39.8_{\tiny \begin{tcolorbox}[bluebox, left=0pt]{\text{$\downarrow$ 1.9}} \end{tcolorbox}}$ & $80.5_{\tiny \begin{tcolorbox}[bluebox, left=0pt]{\text{$\downarrow$ 15.1}} \end{tcolorbox}}$ / $21.1_{\tiny \begin{tcolorbox}[bluebox, left=0pt]{\text{$\downarrow$ 12.2}} \end{tcolorbox}}$ & $66.3_{\tiny \begin{tcolorbox}[bluebox, left=0pt]{\text{$\downarrow$ 0.1}} \end{tcolorbox}}$ & $52.0_{\tiny \begin{tcolorbox}[pinkbox, left=0pt]{\text{$\uparrow$ 0.5}} \end{tcolorbox}}$ \\
    
    \cmark & \cmark & \xmark & $91.9_{\tiny \begin{tcolorbox}[pinkbox, left=0pt]{\text{$\uparrow$ 0.8}} \end{tcolorbox}}$ / $19.4_{\tiny \begin{tcolorbox}[bluebox, left=0pt]{\text{$\downarrow$ 22.3}} \end{tcolorbox}}$ & $95.7_{\tiny \begin{tcolorbox}[pinkbox, left=0pt]{\text{$\uparrow$ 0.1}} \end{tcolorbox}}$ / $00.0_{\tiny \begin{tcolorbox}[bluebox, left=0pt]{\text{$\downarrow$ 33.3}} \end{tcolorbox}}$ & $66.7_{\tiny \begin{tcolorbox}[pinkbox, left=0pt]{\text{$\uparrow$ 0.3}} \end{tcolorbox}}$ & $41.8_{\tiny \begin{tcolorbox}[bluebox, left=0pt]{\text{$\downarrow$ 9.7}} \end{tcolorbox}}$ \\
    
    \cmark & \cmark & \cmark & 91.1 / 41.7 & 95.6 / 33.3 & 66.4 & 51.5 \\
    \bottomrule
    \end{tabular}
    }
    \label{tab:abl_level}
\end{table*}

\subsection{Ablation Study}
\label{abl_results}
We further clarify the contributions of our model through ablation studies, focusing on key components such as the MLLM selection of "Captioner" and "PsyPredictor", the effectiveness of multi-level drawing analysis, and the significance of the object-specific feature generation module and knowledge base integration.

\paragraph{Performance of PICK on Various MLLM selection}
From Table \ref{tab:mllm_performance}, we observe that except for GPT-4o on HTP\_Child, our proposed method consistently achieves higher F1 scores for both positive and negative classes compared to the corresponding baseline models, regardless of the choice of MLLMs for the "captioner" and "PsyPredictor" components across different tasks and datasets. For emotion datasets, our method can achieve better or comparative results compared to corresponding baselines. The complete results with other evaluation metrics can be found in Table
\ref{tab:app_models} and \ref{tab:app_emotion_models} in Appendix \ref{app_results}.

\paragraph{Effectiveness of Hierarchical Framework}
From Table \ref{tab:abl_level}, we observe the following: 1) The Whole and multiple-object levels primarily enhance positive class detection, while the single-object level is crucial for improving negative class performance on HTP datasets.
Specifically, on HTP\_College, PICK (without Whole and multiple-object levels) experiences a 9.7\% drop in positive class performance compared to the full model. 
Furthermore, PICK (without single-object) struggles to detect the negative class, with performance declines of 22.3\% and 33.3\% on the two HTP datasets. On HTP\_Child, without incorporating detailed single-object level analysis, our method fails to detect negative examples, underscoring the necessity of single-object level features for accurate negative class classification.
{This may be attributed to two factors: i) MLLM tend to give a positive prediction due to pretraining biases. At the whole/multi-object level, we directly prompt the MLLM rely on global holistic features and object interactions; ii) Subtle signals—especially those associated with negative cases—are often hidden in the fine details. Our KB and feature extractor enhance the detection of such signals at the single-object level.}
2) By integrating all levels, our proposed method achieves a well-balanced and robust performance across HTP datasets. 
3) On emotion datasets, the whole level analysis is more important than other two levels, shows the significance of whole features for capturing the accurate emotional comprehension. Additional results for other evaluation criteria can be found in Table \ref{tab:app_multi} and \ref{tab:app_emotion_multi} in Appendix \ref{app_results}. 


\begin{table}[]
  \caption{The ablation study on Feature Extraction (FE) Module and Knowledge Base (KB). The results are F1 scores, presented in the format: Positive F1 / Negative F1.}
  \centering
  \setlength{\tabcolsep}{2pt}
  \renewcommand{\arraystretch}{1.1}
  \resizebox{0.45\textwidth}{!}{%
    \begin{tabular}{ccccccc}
    \toprule
    FE & KB & HTP\_College & HTP\_Child & Emotion6 & Artphoto \\
    \midrule
    \xmark & \xmark & 91.5/17.8 & 95.7/0 & 66.7 & 51.6 \\
    \cmark & \xmark & 93.7/30.1 & 95.7/0 & 66.7 & 52.0 \\
    \xmark & \cmark & 90.6/22.8 & 94.4/28.6 & 66.7 & 52.0 \\
    \cmark & \cmark & 91.1/41.7 & 95.6/33.3 & 66.4 & 51.5 \\
    \bottomrule
    \end{tabular}
    }
    \label{tab:abl_KB}
\end{table}

\begin{figure*}[ht]
	\centering
	\includegraphics[width=6.in]{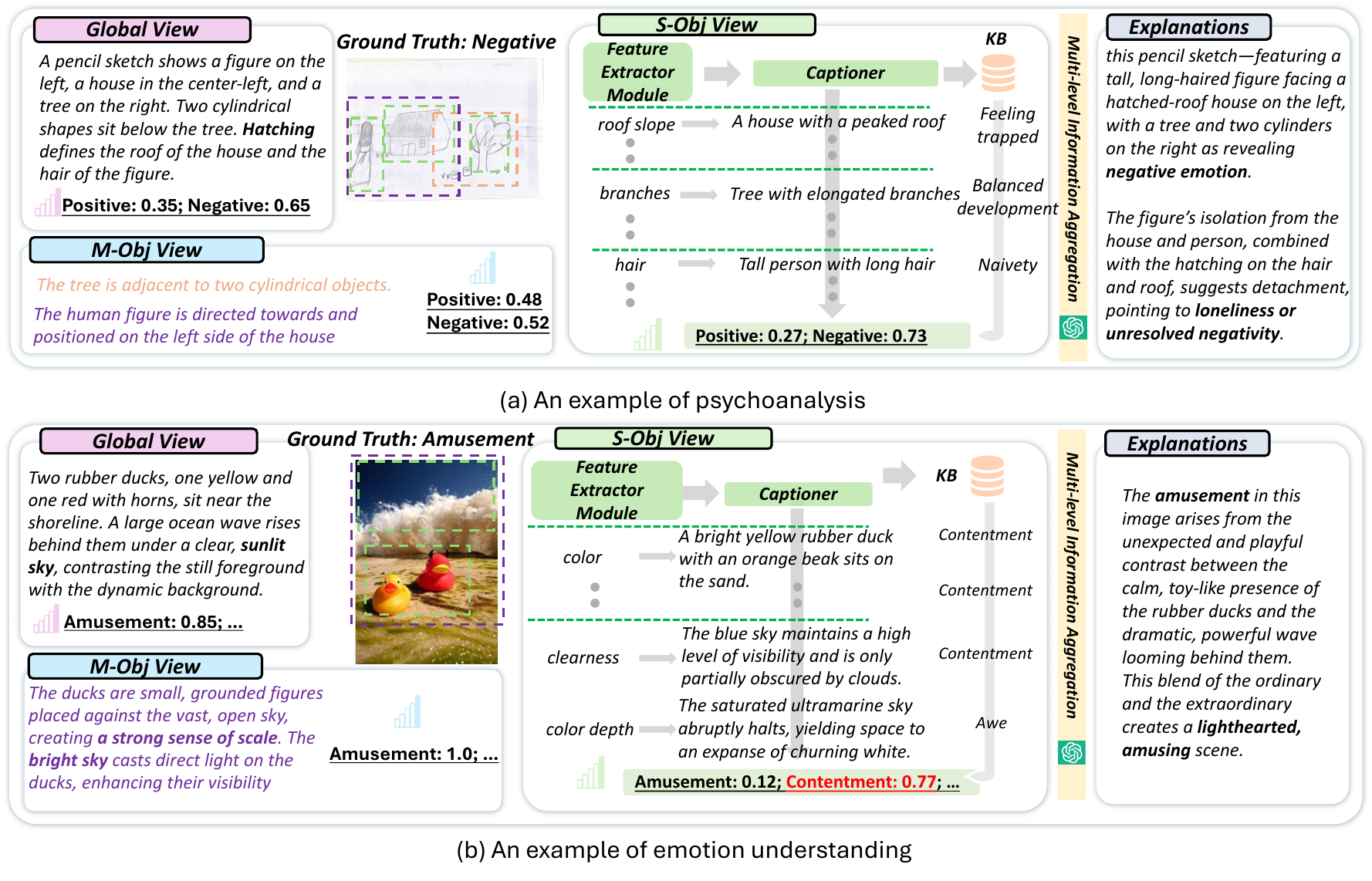}
	\caption{The visualization of PICK with multi-level analysis.
	}
	\label{fig-case1}
\end{figure*}

\paragraph{Effectiveness of Feature Extraction Module and Human Knowledge}
We further investigate the effectiveness of the feature extraction module (FE) and the integration of prediction from human knowledge (KB) $P_{\text{KB}_{i}}^{(S)}$ in Equation \eqref{MLLM_KB}. As shown in Table \ref{tab:abl_KB}, both the FE and KB are critical for the performance of PICK. Notably, their influence is more pronounced for the negative class compared to the positive class. 
In particular, when the FE is removed, the F1 score for the negative class decreases significantly from 41.7 to 22.8 on HTP\_College. This underscores the importance of generating dynamic, object-specific features to capture critical visual cues related to the underlying subtle mental state,while aligning with expert knowledge.
Furthermore, compared to HTP datasets, KB has little influence on emotion datasets. This may be due to the quality of the emotion-related KB, and the fact that emotional expression in images primarily arises from more clear visual cues. More results can be found in Table \ref{tab:app_KB} and \ref{tab:app_emotion_KB} in Appendix \ref{app_results}.

\subsection{Case Study}
We demonstrate the interpretability of the proposed PICK framework by illustrating how it effectively captures critical visual cues and accurately associates them with psychological states, in alignment with expert human interpretations.
Figure \ref{fig-case1} (a) presents the visual description and psychological prediction results from PICK for a negative HTP example from HTP\_College dataset. At the whole level, PICK identifies holistic features such as hatching, which suggest emotional distress, and subsequently predicts a negative psychological tendency. At the single-object level (S-Obj view), the feature extraction module detects object-specific visual attributes—such as the roof slope—which are then translated into descriptive captions (e.g., "A house with a peaked roof") by the Captioner. We observe that according to the Knowledge Base, the description "a house with a peaked roof" is associated with the psychological interpretation of "feeling trapped". Integrating multiple object features, PICK infers a negative tendency at the single-object level as well. Through the integration of multoi-step and multi-level reasoning, PICK generates a coherent and interpretable explanation of the subject’s mental state. 
Figure \ref{fig-case1} (b) illustrates an example of emotion understanding. While the single-object level (S-Obj) predicts a higher probability for contentment, PINK accurately predicts amusement and provides an explanation by aggregating multi-level information. 

Further analyzing a failure case where PICK predicted positive but the ground truth was negative, we observed conflicting cues across levels: the whole-image (0.75/0.25) and single-object (0.66/0.34) levels indicated positive, while the multi-object level (0.25/0.75) indicated negative. This highlights the complexity of psychological tasks and the need for enhanced multi-level integration.
Note that in PICK, the psychological or emotional state is directly predicted at the whole level; the accompanying visual description is provided solely for interpretability and visualization purposes. More examples in Appendix \ref{app_examples}.

\section{Conclusion}
In this work, we explore the potential of Multimodal Large Language Models (MLLMs) in advancing psychological analysis, with a particular focus on the House-Tree-Person (HTP) Test, a widely-used projective drawing technique in clinical psychology. We introduce PICK, a multi-step reasoning framework for drawing-based psychoanalysis. PICK systematically decomposes HTP drawings into semantic meaningful sub-drawings based on object categories, enabling multi-level visual reasoning across the single-object, multi-object, and the whole levels. To support fine-grained interpretation, we further propose a specialized feature extraction module that focuses on detailed visual elements at the object level, facilitating the generation of comprehensive psychological profiles. Our results show that PICK significantly improves MLLMs' ability to assess mental states and emotions, offering a more interpretable approach to analyzing human emotions and cognition through visual expression.

\section*{Ethics Statement}
{Our system is not intended for clinical use, but rather for investigating the capabilities of MLLMs in interpreting subjective visual inputs. Any real-world application—especially in educational or assistive settings—would require additional safeguards and ethical oversight. We caution against misuse in personal or clinical contexts and emphasize the need for professional supervision in any mental health–related use. The dataset used in this study carries inherent limitations, such as binary labeling and expert subjectivity. The HTP datasets were obtained through appropriate channels and used in accordance with applicable data privacy and informed consent regulations.}

\begin{acks}
This work is partially supported by National Natural Science Foundation of China (Grant No. 62372468). The authors sincerely thank the Department of Psychology at Istanbul Bilgi University for providing the data used in this study.
\end{acks}

\clearpage
\bibliographystyle{ACM-Reference-Format}
\bibliography{sample-base}

\clearpage
\appendix

\section{Research Methods}

\subsection{HTP KB construction}
\label{app_KB}

The triplets in the HTP knowledge base are derived from books related to psychoanalysis. The following references were consulted:

\begin{itemize}
    \item \textit{The psychology of graphic images: Seeing, drawing, communicating}
    \item \textit{Using drawings in assessment and therapy: A guide for mental health professionals}
    \item \textit{Illustrated Psychology of Drawing}
    \item \textit{Psychological Drawing: Illustrated Guide to Drawing Psychological Analysis}
    \item \textit{Unlocking the Secrets of Your Personality}
    \item \textit{Tree-Personality Projection Test}
    \item \textit{House-Tree-Person Drawing Projection Test}
    \item \textit{Handbook of Drawing Analysis and Psychotherapy}
    \item \textit{Psychology of Drawing: Pursuing the Meaning Beyond the Picture}
\end{itemize}


\subsection{Prompt examples}
\label{app_prompt}
Table \ref{tab:prompt_summary} presents a set of example prompts.

\begin{table*}[h]
    \caption{Summary of Prompt Functions and Their Templates}
    \centering
    \footnotesize
    \renewcommand{\arraystretch}{1.2}
    \begin{tabular}{p{2cm}|p{3cm}|p{7.5cm}}
        \hline
        \textbf{Module } & \textbf{Prompt Description} & \textbf{Prompt Template} \\
        \hline
        Whole View PsyPredictor & Query the emotional distributions by focusing on the global holistic features &  
        As an emotional psychologist, analyze the all the objects in the sketch drawing and focus on the overall composition, such as layout, use of space, shadow, brushstrokes, symbolism, or other visual characteristics. Determine the underlying emotional distribution of Positive and Negative class. Follow this exact output format: \{Positive: x.xx; Negative: x.xx\} \\
        \hline
        M-Obj View PsyPredictor & Query the emotional distributions by focusing on the interations between objects &  
        As an emotional psychologist, analyze the relationship and interations between all the objects depicted within the green bounding box in the sketch drawing. Determine the underlying emotional distribution of Positive and Negative class. Follow this exact output format: \{Positive: x.xx; Negative: x.xx\} \\
        \hline
        S-Obj View Captioner & Generate a caption related to one attribute of current object & Acting as a emotional psychologist, provide a concise and complete sentence description of the \{object\} depicted within the green bounding box in the sketch drawing. Think carefully and the sentence should focus on the following: \{attribute\}, and should not involve any emotional words. The output structure must be exactly the following: Description: xxx \\
        \hline
        S-Obj View PsyPredictor & Query the emotional distributions by focusing on the image, attribute and caption &  
        As an emotional psychologist, analyze the following: 1. the image, 2. this attribute about the object in the bounding box: \{attribute\} 3. this description based on the image and attribute: \{text\}. Determine the underlying emotional distribution of Positive and Negative class. And assign a confidence score (a float from 0 to 1, where 0 means no confidence and 1 means full confidence) indicating certainty in the emotional interpretation. Follow this exact output format: \{Positive: x.xx; Negative: x.xx\}; Confidence: x.xx \\
        \hline
        Feature Generator (Small MLLM) in Feature Extraction Module & Generate one attribute of current object & Given the provided hand-drawn sketch, focus on the object \{object\} in the green bounding box. Identify one detailed visual attribute of this object that contributes to understanding psychological positive or negative emotions. Avoid mentioning these attributes: \{excluded\_features\} and Color. Be specific and provide ONLY a short phrase. \\
        \hline
    \end{tabular}

    \label{tab:prompt_summary}
\end{table*}

\subsection{Datasets}
\label{app_datasets}

\subsubsection{HTP Datasets}
\textbf{HTP}\_\textbf{College} was collected from the psychological testing center at China University of Petroleum (\cite{pan2022automated}). It includes subjects who are all college students majoring in one of eight disciplines from 2010 to 2021, making it a representative sample of the psychological state of university students. The assessment of the drawings primarily follows the evaluation criteria of the HTP test (\cite{buck1948htp,burns1987kinetic,hongtao2021usage}). After processing, the dataset contains 2093 HTP drawings that have been annotated as either positive or negative based on their applicability. We randomly sampled 20\% of the dataset as the validation set for hyperparameter-tuning, and the remainings are the test set.

\textbf{HTP\_Child}, a dataset containing 257 House-Tree-Person drawings, was provided by the Department of Psychology at Istanbul Bilgi University (\cite{salar2023artificial}). This dataset consists of clinically collected drawings created by children and is analyzed using the child behavior checklist to assess psychological diagnoses. The original annotations include three psychological factors: anxiety (ranging from 0 to 18), depression (ranging from 0 to 26), and externalizing behavior (aggressiveness, ranging from 0 to 70). Following the statistical methods outlined in the checklist, we convert these continuous scores into three binary classification problems using t-scores.

\subsubsection{Emotion Datasets}

\textbf{ArtPhoto} (\cite{machajdik2010affective}) contains 806 artistic photos, eight emotion classes collected from an art sharing site. The ground truth of each image is determined by the user who uploads it. These photos are taken by people who attempt to evoke a certain emotion in the viewer of the photograph through the concious manipulation of the image composition, lighting, colors, etc. The task associated with this dataset is a multi-class emotion recognition problem, which is assigning each image to one of the eight emotion categories.

\textbf{Emotion6} (\cite{emotion6}) was collected from Flickr to generate 1,980 images with six Ekman’s emotion classes  and valence arousal (VA) values. These emotion probability distribution vectors were obtained through a user study, and each image is no longer associated with a single emotion. Therefore, for classification experiments, we assume the emotion class associated with each image is the one with the highest probability. Additionally, we remove one of the emotional class called Surprise as its subjective nature makes it less clearly associated with either positive or negative emotions in the images. 

\subsection{Baseline Model Details}
\label{app_baselines}
\subsubsection{Close Source Models}
\noindent\textbf{GPT-4o} \citep{gpt4o}: 
GPT-4o is a multimodal model capable of processing text, images, and audio, with an estimated size in the hundreds of billions to 1 trillion parameters. Trained on web-scale text, images, and audio, GPT-4o features native multimodal reasoning, multilingual support, and high-speed inference.

\noindent\textbf{Gemini 2.0} \citep{gemini1.5}: 
Gemini 2.0 Flash is a mid-size multimodal model with a Mixture-of-Experts (MoE) architecture, trained on a vast multimodal corpus  with a focus on long-context tasks up to 1 million tokens. 

\subsubsection{Open Source Models}

\noindent\textbf{InternVL2.5} \citep{chen2024internvl}: 
InternVL2 combines a vision Transformer and a language model. It is pretrained on 5M curated multimodal samples, including documents, forms, scientific charts, and medical images. InternVL2.5 ranges from 1B to 108B parameters, pretrained on curated multimodal data including documents, forms, scientific charts, and medical images. It achieves competitive results on specific document-centric tasks, such as DocVQA.

\noindent\textbf{QwenVL2.5} \citep{bai2025qwen2}: 
QwenVL2 is trained on 1.4T tokens, including image-text pairs, OCR data, video, and interleaved documents. With innovations like Naive Dynamic Resolution and Multimodal RoPE, QwenVL2.5 achieves competitive performance on multimodal benchmarks, establishing itself as a leading open-source option.

\noindent\textbf{Idefics3} \citep{Idefics2}: 
Idefics3 combines a Mistral-7B language model with a SigLIP vision encoder. Trained on interleaved web documents, captions, and diagram-text mappings, it supports arbitrary sequences of text and images. Despite its smaller size, it achieves comparable performance to 30B+ models.

All MLLMs are tested using their default settings in the Huggingface environment\footnote{\url{https://huggingface.co/}} with using $2\times$ A100 80G GPUs. 




\subsection{More results}
\label{app_results}

We report the results for HTP\_Child (Anxious) and HTP\_Child (Depressed) in Table \ref{tab:HTP_childAnxious_childDepressed}. Our proposed method significantly improves the diagnosis of negative examples with mental health issues compared to other baseline models.

Tables \ref{tab:app_models} and \ref{tab:app_emotion_models} present the results of the ablation study on MLLM selection across multiple criteria. Our proposed method consistently outperforms the corresponding baselines across different datasets, demonstrating the robustness of our framework to the choice of MLLMs.

Table \ref{tab:app_multi} and \ref{tab:app_emotion_multi} report the results of the ablation study on multi-level analysis on all evaluation critiria across HTP datasets and emotion datsets, respectively. We observe the effectiveness of our hierarchical framework in capturing nuanced psychoanalytical features, as evidenced by consistent improvements across various evaluation criteria in both HTP and emotion datasets. The results highlight the importance of multi-level analysis in refining the interpretability and accuracy of psychological assessments.

Tables \ref{tab:app_KB} and \ref{tab:app_emotion_KB} present the results of the object-specific feature generation module and the integration of the knowledge base. Our findings indicate that incorporating these components helps improve the diagnosis of negative examples in HTP datasets. However, the model achieves similar performance to the full framework even when either the object-specific feature generation module or the knowledge base integration is removed.

\begin{table*}[]
  \caption{Results on HTP\_Child(Anxious) and HTP\_Child(Depressed) datasets.}
  \centering
  \setlength{\tabcolsep}{4pt}
  \renewcommand{\arraystretch}{1.2}
  \resizebox{0.85\textwidth}{!}{%
  \begin{tabular}{l
      *{7}{c}
      *{7}{c} }
    \toprule
    \multirow{3}{*}{\textbf{Method}} 
      & \multicolumn{7}{c}{\textbf{HTP\_Child(Anxious)}} 
      & \multicolumn{7}{c}{\textbf{HTP\_Child(Depressed)}} \\
    \cmidrule(lr){2-8} \cmidrule(l){9-15}
      & \multirow{2}{*}{Acc} 
      & \multicolumn{3}{c}{Positive} & \multicolumn{3}{c}{Negative} 
      & \multirow{2}{*}{Acc} 
      & \multicolumn{3}{c}{Positive} & \multicolumn{3}{c}{Negative} \\
    \cmidrule(lr){3-5} \cmidrule(lr){6-8} \cmidrule(lr){10-12} \cmidrule(l){13-15}
      &   & F1 & Prec & Rec & F1 & Prec & Rec 
      &   & F1 & Prec & Rec & F1 & Prec & Rec \\
    \midrule
    GPT-4o                & 83.3  & 90.9 & 85.1 & 97.6 & 0    & 0    & 0    & 87.5  & 93.3 & 87.5 & 100  & 0    & 0    & 0    \\
    Gemini-2.0-Flash      & 64.6  & 76.1 & 90.0 & 65.9 & 32.0 & 22.2 & 57.1 & 64.6  & 77.9 & 85.7 & 71.4 & 10.5 & 7.7  & 16.7 \\
    Qwen2.5-VL-7B         & 72.9  & 84.3 & 83.3 & 85.4 & 0    & 0    & 0    & 68.8  & 81.5 & 84.6 & 78.6 & 0    & 0    & 0    \\
    Qwen2.5-VL-72B        & 81.3  & 89.7 & 84.8 & 95.1 & 0    & 0    & 0    & 85.4  & 92.1 & 87.2 & 97.6 & 0    & 0    & 0    \\
    Idefics3-8B           & 64.6  & 77.3 & 85.3 & 70.7 & 19.0 & 14.3 & 28.6 & 62.5  & 76.3 & 85.3 & 69.0 & 10.0 & 7.1  & 16.7 \\
    InternVL2.5-8B        & 62.5  & 76.3 & 82.9 & 70.7 & 10.0 & 7.7 & 14.3 & 62.5  & 75.7 & 87.5 & 66.7 & 18.2 & 12.5 & 33.3 \\
    InternVL2.5-78B       & 64.6  & 77.3 & 85.3 & 70.7 & 19.0 & 14.3 & 28.6 & 87.5  & 93.3 & 87.5 & 100  & 0    & 0    & 0    \\
    \hline
    \rowcolor{gray!30} {PICK (Based on Gemini)}
                          & 72.9  & 82.2 & {93.8} & 73.2 & {43.5} & {31.3} & {71.4} & 70.8 & 80.0 & {100}  & 66.7 & {46.2} & {30.0} & {100} \\
    \bottomrule
  \end{tabular}%
  }
  \label{tab:HTP_childAnxious_childDepressed}
\end{table*}

\begin{table*}[]
  \caption{The ablation study of MLLM selections on HTP\_College and HTP\_Child(Aggressive) datasets.}
  \centering
  \setlength{\tabcolsep}{5pt}
  \renewcommand{\arraystretch}{1.1}
  \resizebox{0.85\textwidth}{!}{%
  \begin{tabular}{l
      *{7}{c}
      *{7}{c} }
    \toprule
    \multirow{3}{*}{\textbf{Method}} 
      & \multicolumn{7}{c}{\textbf{HTP\_College}} 
      & \multicolumn{7}{c}{\textbf{HTP\_Child(Aggressive)}} \\
    \cmidrule(lr){2-8} \cmidrule(l){9-15}
      & \multirow{2}{*}{Acc} 
      & \multicolumn{3}{c}{Positive} & \multicolumn{3}{c}{Negative} 
      & \multirow{2}{*}{Acc} 
      & \multicolumn{3}{c}{Positive} & \multicolumn{3}{c}{Negative} \\
    \cmidrule(lr){3-5} \cmidrule(lr){6-8} \cmidrule(lr){10-12} \cmidrule(l){13-15}
      &   & F1 & Prec & Rec & F1 & Prec & Rec 
      &   & F1 & Prec & Rec & F1 & Prec & Rec \\
    \midrule
    GPT-4o                & 68.9  & 80.0 & 92.0 & 70.8 & 29.5 & 20.2 & 54.6  & 91.7  & 95.7 & 91.7 & 100  & 0    & 0    & 0    \\
    Ours      & 81.1  & 88.8 & 93.0 & 85.0 & 40.1 & 32.3 & 52.8  & 91.7  & 95.7 & 91.7 & 100  & 0    & 0    & 0 \\
    \midrule
    Qwen2.5-VL-7B         & 66.9  & 78.2 & 92.9 & 67.6 & 30.9 & 20.6 & 62.0  & 85.4  & 92.0 & 93.0 & 90.9 & 22.2 & 20.0 & 25.0 \\
    Ours           & 68.2  & 78.9 & 94.6 & 67.7 & 34.9 & 23.1 & 71.6  & 81.3  & 89.2 & 94.9 & 84.1 & 30.8 & 22.2 & 50.0 \\
    \midrule
    InternVL2.5-8B        & 69.0  & 80.5 & 90.1 & 72.8 & 24.1 & 17.1 & 41.2  & 85.4  & 92.0 & 93.0 & 90.9 & 22.2 & 20.0  & 25.0 \\
    Ours      & 77.9  & 87.0 & 90.1 & 84.2 & 25.5 & 21.4 & 31.6  & 87.5  & 93.0 & 95.2 & 90.9 & 40.0 & 33.3 & 50.0 \\
    \bottomrule
  \end{tabular}%
  }
  \label{tab:app_models}
  
\end{table*}

\begin{table*}[]
  \caption{The ablation study of MLLM selection on Emotion datasets.}
  \centering
  \setlength{\tabcolsep}{4pt}
  \renewcommand{\arraystretch}{1.2}
  
  \resizebox{0.7\textwidth}{!}{%
  \begin{tabular}{lcccccccc}
    \toprule
    \multirow{2}{*}{\textbf{Method}}  
    & \multicolumn{4}{c}{\textbf{Emotion6}} 
    & \multicolumn{4}{c}{\textbf{ArtPhoto}} \\
    \cmidrule(r){2-5} \cmidrule(l){6-9}
    & {Acc} & {F1} & {Prec} & {Rec} 
    & {Acc} & {F1} & {Prec} & {Rec} \\
    \midrule
    GPT-4o & 68.2 & 65.2 & 71.6 & 68.3 & 47.4 & 45.7 & 58.7 & 45.5 \\
    Ours & 68.6 & 65.1 & 72.7 & 68.6 & 47.5 & 49.1 & 59.0 & 47.5 \\
    \midrule
    Qwen2.5-VL-7B & 68.2 & 62.4 & 73.5 & 68.2 & 43.9 & 42.7 & 52.6 & 43.3 \\
    Ours & 66.1 & 61.1 & 71.2 & 66.1 & 42.6 & 43.2 & 54.0 & 42.6 \\
    \midrule
    InternVL2.5-8B & 64.5 & 61.3 & 65.8 & 64.5 & 45.2 & 43.8 & 49.2 & 42.4 \\
    Ours & 67.4 & 63.3 & 69.3 & 67.4 & 45.9 & 46.9 & 53.2 & 46.9 \\
  \bottomrule
  \end{tabular}
  }
  \label{tab:app_emotion_models}
\end{table*}

\begin{table*}[]

  \caption{The ablation study of hierarchical analysis on HTP\_College and HTP\_Child(Aggressive) datasets. w/o is without.}
  \centering
  \setlength{\tabcolsep}{5pt}
  \renewcommand{\arraystretch}{1.1}
  \resizebox{0.8\textwidth}{!}{%
  \begin{tabular}{l
      *{7}{c}
      *{7}{c} }
    \toprule
    \multirow{3}{*}{\textbf{Method}} 
      & \multicolumn{7}{c}{\textbf{HTP\_College}} 
      & \multicolumn{7}{c}{\textbf{HTP\_Child(Aggressive)}} \\
    \cmidrule(lr){2-8} \cmidrule(l){9-15}
      & \multirow{2}{*}{Acc} 
      & \multicolumn{3}{c}{Positive} & \multicolumn{3}{c}{Negative} 
      & \multirow{2}{*}{Acc} 
      & \multicolumn{3}{c}{Positive} & \multicolumn{3}{c}{Negative} \\
    \cmidrule(lr){3-5} \cmidrule(lr){6-8} \cmidrule(lr){10-12} \cmidrule(l){13-15}
      &   & F1 & Prec & Rec & F1 & Prec & Rec 
      &   & F1 & Prec & Rec & F1 & Prec & Rec \\
    \midrule
    Only Whole         & 85.4  & 90.8 & 89.4 & 92.2 & 22.2 & 25.5 & 19.6  & 85.4  & 92.0 & 93.0 & 90.9 & 22.2 & 20.0 & 25.0 \\
    Only M-Obj         & 84.5  & 91.5 & 89.0 & 94.1 & 17.8 & 24.3 & 14.0  & 91.7  & 95.7 & 91.7 & 100 & 0 & 0 & 0 \\
    Only S-Obj         & 71.3  & 81.4 & 94.5 & 71.6 & 36.5 & 24.8 & 69.1  & 50.0  & 63.6 & 95.5 & 47.7 & 20.0 & 11.5 & 75.0 \\
    Full Model w/o Whole         & 84.0  & 90.7 & 93.2 & 88.3 & 44.2 & 37.9 & 52.8  & 89.6  & 89.6 & 91.5 & 97.7 & 0 & 0 & 0 \\
    Full Model w/o M-Obj           & 77.2 & 85.9 & 94.1 & 79.1 & 39.8 & 29.0 & 63.2  & 68.8  & 80.5 & 93.9 & 70.5 & 21.1 & 13.3 & 50.0 \\
    Full Model w/o S-Obj 
                          & 85.3 & 91.9 & 89.9 & 91.2 & 19.4 & 28.2 & 14.8  & 91.7  & 95.7 & 91.7 & 100  & 0    & 0    & 0    \\
    \textbf{Full Model} 
                          & {84.7}  & {91.1} & 92.5 & {89.9} & {41.7} & {38.2} & 46.0  & {91.7}  & {95.6} & 93.5 & 97.7 & {33.3} & {50.0} & 25.0 \\
    \bottomrule
  \end{tabular}%
  }
  \label{tab:app_multi}
  
\end{table*}

\begin{table*}[]

  \caption{The ablation study of hierarchical analysis on Emotion datasets. w/o is without.}
  \centering
  \setlength{\tabcolsep}{4pt}
  \renewcommand{\arraystretch}{1.2}

  \resizebox{0.7\textwidth}{!}{%
  \begin{tabular}{lcccccccc}
    \toprule
    \multirow{2}{*}{\textbf{Method}}  
    & \multicolumn{4}{c}{\textbf{Emotion6}} 
    & \multicolumn{4}{c}{\textbf{ArtPhoto}} \\
    \cmidrule(r){2-5} \cmidrule(l){6-9}
    & {Acc} & {F1} & {Prec} & {Rec} 
    & {Acc} & {F1} & {Prec} & {Rec} \\
    Only Whole  & 70.6 & 66.7 & 73.1 & 70.6 & 49.9 & 51.4 & 58.1 & 49.9 \\
    Only M-Obj & 69.8 & 66.4 & 71.7 & 69.8 & 49.4 & 50.9 & 57.4 & 49.4 \\
    Only S-Obj & 62.1 & 57.8 & 64.2 & 62.1 & 40.1 & 42.2 & 53.5 & 40.1 \\
    Full Model w/o Whole & 66.7 & 62.9 & 69.2 & 66.7 & 43.2 & 45.3 & 53.9 & 43.2 \\
    Full Model w/o M-Obj & 70.2 & 66.3 & 72.7 & 70.2 & 50.6 & 52.0 & 58.6 & 50.6 \\
    Full Model w/o S-Obj & 70.6 & 66.7 & 73.1 & 70.6 & 44.5 & 41.8 & 49.8 & 41.8 \\
    Full Model & 70.3 & 66.4 & 73.0 & 70.3 & 50.1 & 51.5 & 58.1 & 50.1 \\
  \bottomrule
  \end{tabular}
  }
  \label{tab:app_emotion_multi}
\end{table*}

\begin{table*}[]
  \caption{The ablation study of feature extraction module and KB on HTP\_College and HTP\_Child(Aggressive) datasets. w/o is without.}
  \centering
  \setlength{\tabcolsep}{5pt}
  \renewcommand{\arraystretch}{1.1}
  \resizebox{0.85\textwidth}{!}{%
  \begin{tabular}{l
      *{7}{c}
      *{7}{c} }
    \toprule
    \multirow{3}{*}{\textbf{Method}} 
      & \multicolumn{7}{c}{\textbf{HTP\_College}} 
      & \multicolumn{7}{c}{\textbf{HTP\_Child(Aggressive)}} \\
    \cmidrule(lr){2-8} \cmidrule(l){9-15}
      & \multirow{2}{*}{Acc} 
      & \multicolumn{3}{c}{Positive} & \multicolumn{3}{c}{Negative} 
      & \multirow{2}{*}{Acc} 
      & \multicolumn{3}{c}{Positive} & \multicolumn{3}{c}{Negative} \\
    \cmidrule(lr){3-5} \cmidrule(lr){6-8} \cmidrule(lr){10-12} \cmidrule(l){13-15}
      &   & F1 & Prec & Rec & F1 & Prec & Rec 
      &   & F1 & Prec & Rec & F1 & Prec & Rec \\
    \midrule
    Full Model w/o FE-Module and KB         & 84.5  & 91.5 & 89.0 & 94.1 & 17.8 & 24.3 & 14.0  & 91.7  & 95.7 & 91.7 & 100 & 0 & 0 & 0 \\
    Full Model w/o KB           & 88.4 & 93.7 & 90.1 & 97.6 & 30.1 & 54.2 & 20.8  & 91.7  & 95.7 & 91.7 & 100 & 0 & 0 & 0 \\
    Full Model w/o FE-Module        & 83.2 & 90.6 & 89.5 & 91.6 & 22.8 & 25.2 & 20.8  & 89.6  & 94.4 & 93.3 & 95.5 & 28.6 & 33.3  & \textbf{25.0} \\
    Full Model                                & {84.7}  & {91.1} & \textbf{92.5} & {89.9} & \textbf{41.7} & \textbf{38.2} & \textbf{46.0}  & \textbf{91.7}  & {95.6} & \textbf{93.5} & 97.7 & \textbf{33.3} & \textbf{50.0} & \textbf{25.0} \\
    \bottomrule
  \end{tabular}%
  }
  \label{tab:app_KB}
  
\end{table*}

\begin{table*}[]
  \caption{The ablation study of feature extraction module and KB on Emotion datasets. w/o is without.}
  \centering
  \setlength{\tabcolsep}{4pt}
  \renewcommand{\arraystretch}{1.2}
  \resizebox{0.7\textwidth}{!}{%
  \begin{tabular}{lcccccccc}
    \toprule
    \multirow{2}{*}{\textbf{Method}}  
    & \multicolumn{4}{c}{\textbf{Emotion6}} 
    & \multicolumn{4}{c}{\textbf{ArtPhoto}} \\
    \cmidrule(r){2-5} \cmidrule(l){6-9}
    & {Acc} & {F1} & {Prec} & {Rec} 
    & {Acc} & {F1} & {Prec} & {Rec} \\
    Full Model w/o FE-Module and KB & 70.6 & 66.7 & 73.0 & 70.6 & 50.2 & 51.6 & 58.2 & 50.2 \\
    Full Model w/o KB & 70.6 & 66.7 & 73.1 & 70.6 &  50.5 & 52.0 & 58.7 & 52.0 \\
    Full Model w/o FE-Module & 70.6 & 66.7 & 72.7 & 70.6 & 50.5 & 52.0 & 58.4 & 50.5 \\
    Full Model & 70.3 & 66.4 & 73.0 & 70.3 & 50.1 & 51.5 & 58.1 & 50.1 \\
  \bottomrule 
  \end{tabular}
  }
  \label{tab:app_emotion_KB}
\end{table*}

\subsection{More Examples}
\label{app_examples}

\begin{figure*}[]
	\centering
	\includegraphics[width=5.3in]{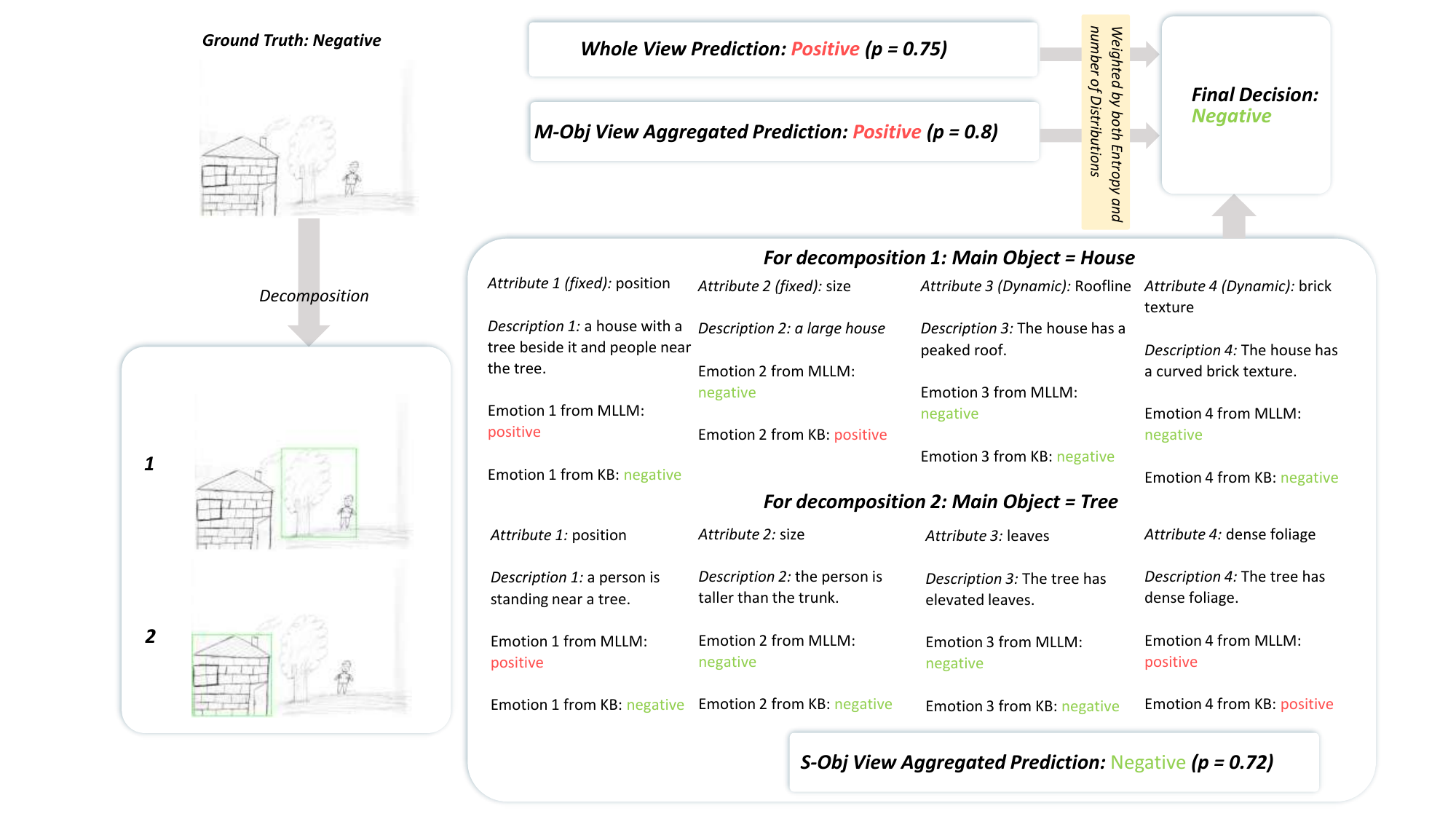}
	\caption{The visualization of PICK with multi-level analysis on an negative example correctly predicted as negative.
	}
	\label{fig-casem1}
\end{figure*}

We observe that in Figure \ref{fig-casem1}, our method effectively captures key psychoanalytical cues that indicate underlying psychological distress, highlighting its ability to distinguish subtle features associated with mental health concerns.

\begin{figure*}[]
	\centering
	\includegraphics[width=5.3in]{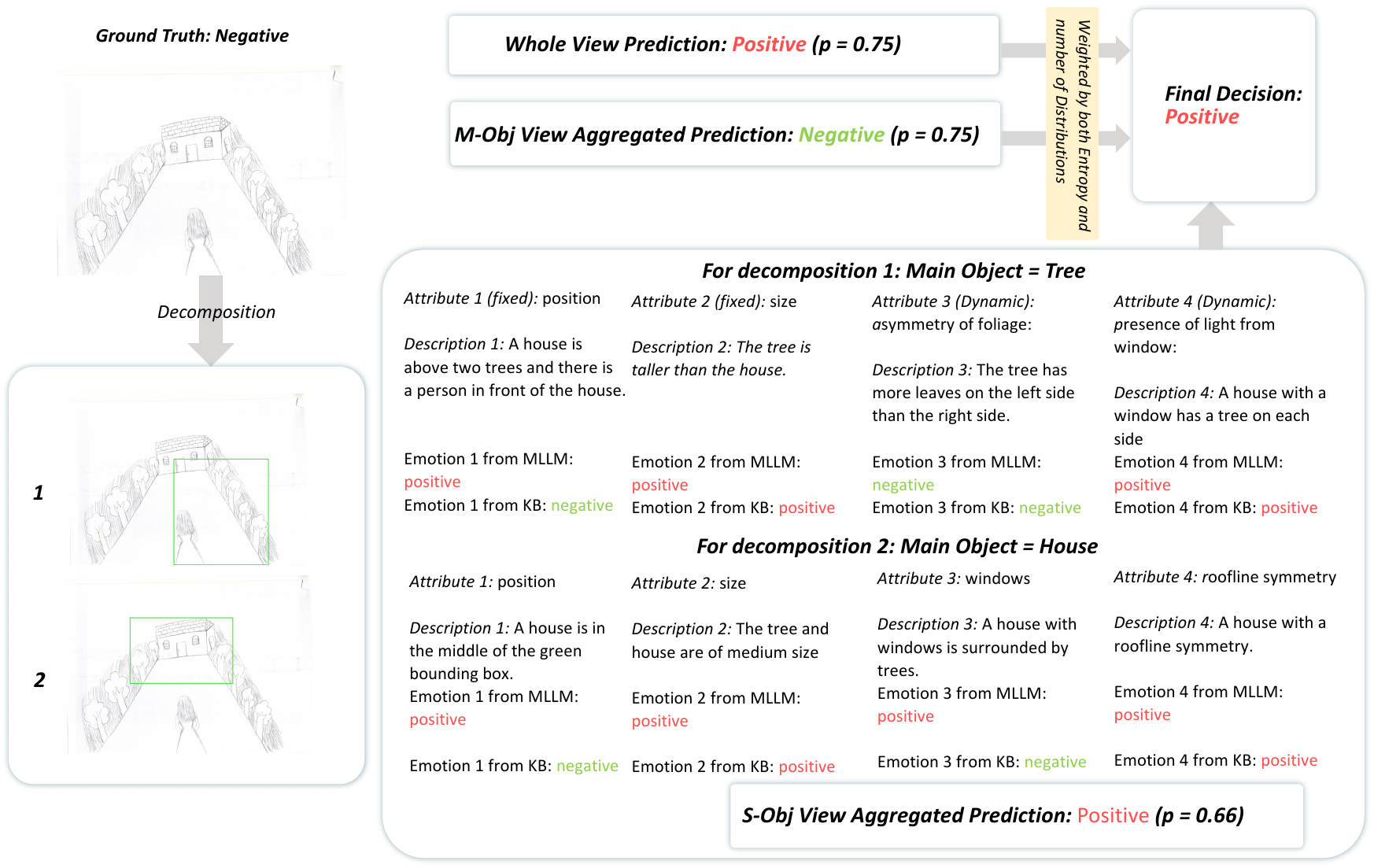}
	\caption{The visualization of PICK with multi-level analysis on an negative example incorrectly classified as positive.
	}
	\label{fig-casem2}
\end{figure*}

After analyzing PICK’s failure on a drawing, Figure \ref{fig-casem2}, with a negative ground truth but a positive prediction, we observe that PICK predicts positive at the whole-level (0.75 positive, 0.25 negative), negative at the multi-object level (0.25 positive, 0.75 negative), and positive at the single-object level (0.66 positive, 0.34 negative). This reveals the complexity of the object’s psychological state, with conflicting cues across different levels of analysis. The failure likely arises from PICK’s inability to reconcile these contradictory signals into a consistent prediction, highlighting the need for better integration methods to balance multi-level information. In the revised version, we will include failure cases to illustrate the limitations of the proposed method and discuss potential improvements.



\end{document}